\documentclass[runningheads]{llncs}
\usepackage{graphicx}
\usepackage{subfigure}

\begin{document}

\title{Synthesizing New Retinal Symptom Images by Multiple Generative Models} 
\titlerunning{Synthesizing New Retinal Symptom Images} 

\author{Yi-Chieh Liu\inst{1}\and
Hao-Hsiang Yang\inst{1}\and
C.-H. Huck Yang\inst{2,3}\and
Jia-Hong Huang\inst{1,2}\and
Meng Tian\inst{4}\and
Hiromasa Morikawa\inst{2}\and
Yi-Chang James Tsai\inst{3}\and
Jesper Tegn\`er\inst{2, 5}}

\authorrunning{Y.-C. Liu et al.}

\institute{
National Taiwan University, Taiwan \email{\{b01310048,r05921014\}@ntu.edu.tw} \and 
Biological and Environmental Sciences and Engineering Division,
Computer, Electrical and Mathematical Sciences and Engineering
Division, King Abdullah University of Science and Technology, Thuwal, Saudi Arabia \email{\{chao-han.yang, jiahong.huang, hiromasa.morikawa, jesper.tegner\}}@kaust.edu.sa\\ \and
Georgia Institute of Technology, GA, USA\\ \email{james.tsai@ce.gatech.edu}\\\and 
Department of Ophthalmology, Bern University Hospital, Bern, Switzerland \email{tianmeng1231@gmail.com} \and
Unit of Computational Medicine, Center for Molecular Medicine,Department of Medicine, Karolinska Institutet, Stockholm, Sweden}

%




\maketitle

\begin{abstract}
Age-Related Macular Degeneration (AMD) is an asymptomatic retinal disease which may result in loss of vision. There is limited access to high-quality relevant retinal images and poor understanding of the features defining sub-classes of this disease. Motivated by recent advances in machine learning we specifically explore the potential of generative modeling, using Generative Adversarial Networks (GANs) and style transferring, to facilitate clinical diagnosis and disease understanding by feature extraction. We design an analytic pipeline which first generates synthetic retinal images from clinical images; a subsequent verification step is applied. In the synthesizing step we merge GANs (DCGANs and WGANs architectures) and style transferring for the image generation, whereas the verified step controls the accuracy of the generated images. We find that the generated images contain sufficient pathological details to facilitate ophthalmologists' task of disease classification and in discovery of disease relevant features. In particular, our system predicts the drusen and geographic atrophy sub-classes of AMD. Furthermore, the performance using CFP images for GANs outperforms the classification based on using only the original clinical dataset. Our results are evaluated using existing classifier of retinal diseases and class activated maps, supporting the predictive power of the synthetic images and their utility for feature extraction. Our code examples are available online. \footnote{https://github.com/huckiyang/EyeNet-GANs} 
\end{abstract}
\section{Introduction}
As a rule, it is challenging to automatically diagnose retinal diseases from images, partly because of the difficulty of acquiring public data with a sufficient number of annotated images due to concerns of personal privacy. Meanwhile, different ophthalmologists may provide conflicting judgments about identical images; therefore, it can be arduous to reach consensus about a diagnosis.Thus, it is clear that a larger number of retinal images collected from a system with provided unbiased feature detection would be beneficial for ophthalmologists’ clinical practice. 
\par
Generative models, such as Generative Adversarial Networks \cite{goodfellow2014generative} (GANs), and style transferring \cite{luan2017deep} techniques, have achieved impressive results for generating sharp and realistic images. Therefore, these two methods are used to synthesize the disease images from healthy retinal images and diseased ones. Synthesized images not only impose high-level symptom features to the original ones but help ophthalmologists build the understanding of related diseases. The definition of image synthesis in \cite{NIPS2017_6642} is seen as an image reconstruction process coupled with feature transformation. The synthesized part is responsible for inverting features back to the color space and the feature transformation matches certain statistics of a original image to a generated image \cite{salehinejad2018generalization}. 
\par
We consider images with Age-Related Macular Degeneration (AMD) as an asymptomatic retinal disease and the leading cause of irreversible visual loss among the aged population. Despite the advances of therapeutics, there is still no satisfactory treatment. It raises the issue that diagnosing AMD from its early stage and having proper managing it properly are more important than ever. The development of AMD is classified as several stages that can be discerned by two explicit symptoms, drusen and Geographic Atrophy(GA). Drusen are one of the earliest clinical indications of AMD, which appears as focal, with yellow excrescences deep in the retina with extra-cellular deposits located beneath the retinal pigment, epithelium, and Bruch’s membrane; the number, size and distribution of these deposits is highly variable. GA, symptomatic of a more advanced stage of AMD, is described as a well-demarcated area of decreased retinal thickness. Such areas have relative changes in color compared to surroundings allowing an increased visualization of the underlying choroidal vessels. The phenomenon is that less intense and more diffuse hyperfluorescence in which pigment clumping sometimes forms a microreticular pattern, is demonstrated \cite{klein2007fifteen} \cite{green1985pathologic} \cite{gheorghe2015age}. To sum up, two symptoms (drusen and GA) are established clinical hallmarks of AMD. Drusen size and confluency have been historically associated with the progression of AMD, which also contributes to the development of GA. Our chief objective is to generate images equipped with a sufficient number of pathological features to capture the two different stages of AMD. \par
The contribution of this paper consists of two parts. First,style transferring, WGANs and DCGANs are used to build a new artificial neural network as the framework for the generation of synthetic pathologically relevant but detailed images. Second, after new images are obtained and diagnosed by ophthalmologists, we use Class Activation Maps (CAMs) \cite{zhou2015cnnlocalization} to locate the advanced features within the generated images. Finally, the EyeNet \cite{yang2018novel} is used to classify generated images according to the established labelling of diseases. 
\par
The paper is organized into five sections. Following this introduction,in the second section, we survey related work. In the third section, we present our analytic pipeline, including an account of how we fuse DCGANs, WGANs, style transferring, EyeNet, and CAMs. In the fourth section, we present and discuss computational experimental results. In the last section, we summarize conclusions and outline prospects the future work.

\section{Related Work}
Below, we survey previous work on GANs, from which we benefit, synthetic image generation, and computational retinal disease methods. 

\noindent\textbf{2.1 GANs}

Since the pioneering formulation of GANs \cite{goodfellow2014generative}, there have been numerous studies of how to formulate the optimization problem of balancing on the one hand the training of a generative network \textbf{G} producing realistic synthetic samples, and on the other, a discriminator network \textbf{D} that distinguishes between real and synthetic (generated) data. We adopt an adversarial loss 

\begin{equation}
    min_Gmax_DL_{GANS}= E_{x\sim p_{data(x)}}[logD(x)]+E_{x\sim p_{prior(z)}}[log(1-D(z))]
\label{eq:gan}
\end{equation}

Yet, a major issue has been the stability and convergence of training a GAN. Recent work \cite{arjovsky2017wasserstein} demonstrated improved stability when using a Kantorovich-Rubinstein metric, which we have adopted in our training of the GAN for retinal images. Rapid advances have demonstrated that GANs generate realistic images, with a rich number of features. For example, GANs have been successfully applied for face generation \cite{pumarola2018GANimation}, indoor scene reconstruction \cite{fan2017point} and person re-identification \cite{qian2017pose}.
Here we benefit from recent progress with GANs to generate new synthetic retinal disease images using both Deep Convolutional Generative Adversarial Networks (DCGANs) \cite{radford2015unsupervised} and Wasserstein GANs (WGANs) \cite{arjovsky2017wasserstein}. These architectures utilize a convolutional decoder, and DCGAN enables the employment of large GANs using Convolutional Neural Networks (CNNs), resulting in stable training across various datasets. Finally, our use of WGANs improves the stability of learning,thereby avoiding known challenges such as model collapse \cite{arjovsky2017wasserstein}.

\noindent\textbf{2.2 Generation of Synthetic Images}

Recently, researchers have used convolutional neural networks to generate images \cite{gatys2016image} with different given style. The method makes use of a pre-trained network to optimize the image and its features. However, this method operates as a global optimization; therefore, generated image exhibit distortions and detailed parts cannot be presented on the transferred images. Meeting this challenge, recent work \cite{luan2017deep} accomplished realistic image generation and style transferring. On the other hand, in \cite{radford2015unsupervised}, authors combined a convolutional neural network and GANs to generate new images, thus mitigating the impact of a limited number of features of pathological relevance in original images. While their work clearly improved the state-of-the-art methods, the technique may generate poor image samples or fails to converge. To ensure convergence and quality of generated images, we deploy a closed form solution for style transferring \cite{li2018closed}.

\noindent\textbf{2.3 Computational Retinal Disease Methods}

There is a challenge to build large high-quality medical databases despite massive investment in, for example, data collection, labeling, and data augmentation. Exceptions include the recently released ChestXray14 \cite{wang2017chestx} dataset which contains 112,120 frontal-view chest radiographs with up to 14 thoracic pathological labels. Yet, in contrast, for retinal research, the DRIVE dataset \cite{staal2004ridge}, which contains only 40 retina images, has long been a standard. However, recently the Retina Image Bank (RIB) \cite{rib}, containing a large number of different kinds of retinal images, is truly an enabler for the kind of work presented in our paper. Despite this, we still need techniques to augment such databases due to various challenges, such as the limited amount of annotation, thus effectively transforming a small dataset with low diversity into one that approximates the underlying data distribution. For example, in \cite{beers2018high} and \cite{guibas2017synthetic}, GANs were used to generate a variety of retinal images and targeting control (healthy) images. Using the Retinal Image Bank, we aim to generate new retinal images that have a sufficient number of pathological details so that we have abundant and useful retinal images to train and build robust classifier. In \cite{kaur2015segmentation}, authors propose a method that implements automated segmentation of retina to facilitate the detection of disease. Article \cite{yang2018novel} uses the whole retinal images in\cite{rib} to train the classifier, which can discern multiple diseases with the extraction of visual traits. Our work depends on having a pre-trained network to test the quality of generated images and uses CAMs to present symptoms identified by the classifier.  

\section{Methodology}
In this section, we describe different methods in our proposed pipeline. For generative models, GANs and style transferring based networks are discussed. For verification, we elaborate EyeNet and CAMs.

\noindent\textbf{3.1 Style Transferring}

The input contains two images: a content and a style image and pre-trained CNNs; the output is the synthesized image. In our case, the content image means the disease image with pathological details; the style image represents the healthy retinal image. When it comes to the existing style transferring methods, even though the style is changed, the content of the image can be seen in the new image. Thus, we expect generated images with pathological details, so content images are seen as disease images. For each image, the output from the CNNs classifier obtains various level features from many convolutional layers. Generated images preserve the original semantic content from the content image but look like a style image. For the content and the style part, loss functions that are computed from the similarity of images from convolutional layers can be defined; style transferring becomes an optimization problem when the optimal image is obtained with the least loss. The pixels in the image can be computed iteratively by gradient descent.

\noindent\textbf{3.2 DCGANs and WGANs for image generation}

Although original GANs provide an intriguing algorithm with surprising results, the instability is what we concern about when it comes to medical applications, which requires precision and detailed images for diagnosis. To improve the quality of generated images, we chose DCGANs and WGANs to establish our generative model. In this part, with a random initialized parameter, we build a generator of retinal diseases while the improving discriminator. For a specific symptom, generated images contain similar optical traits. Furthermore, high dimensional neural networks for computer vision sometimes materialize higher forms of neglected visual features. Therefore, generated retinal images not only become the aid of diagnosis and strategy to explore diseases, but also provide diverse computer training data.

\noindent\textbf{3.3 Class Activation Maps}

The class activation maps (CAMs) in \cite{zhou2015cnnlocalization} provide a method that localizes features on images. From localized features, the performance of the generated image can be evaluated and observed. As discussed in Section 3.1, convolutional layers of CNNs are used to extract the visual feature of images. Through this method, not only the similarity of images is tested with high-level disease features, but a series of pathological details is built.

\noindent\textbf{3.4 EyeNet}

Besides CAMs, EyeNet as proposed in \cite{yang2018novel} is used to evaluate the correctness of generated images. In \cite{yang2018novel}, the authors trained a network that classifies different retinal diseases; 52 kinds of retinal diseases are labeled and classified. Proposed methods \cite{yang2018auto} include three frameworks: U-net, SVM and ResNet50; predicton by ResNet50 performs best. Therefore, ResNet50 is modified so that the generated images also can be classified to make sure of their correctness.

\noindent\textbf{3.5 Pipeline}

\begin{figure*}[h]
  \centering
\includegraphics[width=\linewidth]{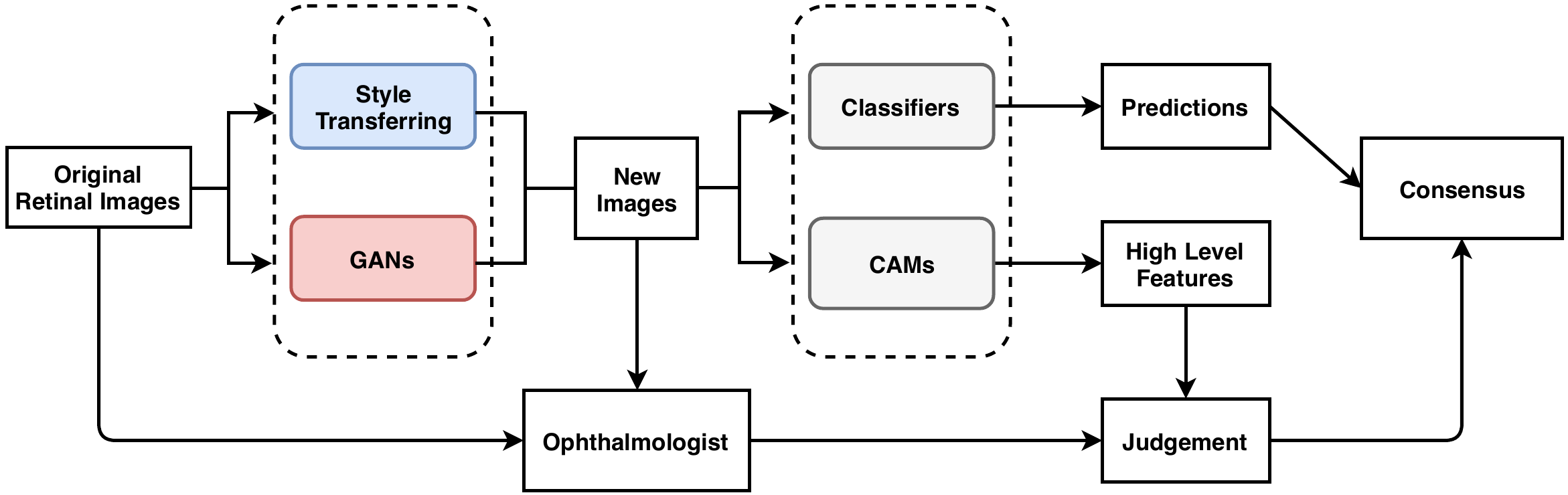}
    \caption{{Proposed pipeline to improve retinal diagnosis efficiency by combining GANs and EyeNet to assist ophthalmologists and doctors. EyeNet is used to check the performance. Finally, observation of similarities among some diseases are analyzed and described.}}
\label{fig:figure1}
\end{figure*}

We propose a pipeline structure in Fig. \ref{fig:figure1}. Initially, with feature extraction by style transferring and GANs, more images are generated. In order to verify the correctness, CAMs and EyeNet are used to compute the high level visual features and predict the diseases, respectively. Results from the CAMs present pathological details. Moreover, generated images can be applied to feed to other classifier to train the more accurate classifier. All researches benefits not only the newly trained network, but also ophthalmologists.
Original retinal images give doctors an initial diagnosis, and the generated images provide them more clues. CAMs help ophthalmologists judge accurately, and they can reach the consensus with EyeNet. To sum up, our pipeline improves the efficiency and accuracy of the medical system and contributes to researchers.

\section{Experiments}
In this section, we describe the implementation details and experiments we conduct to validate our proposed methods. Initially, the data collection and setup of experiments are emphasized. And generating images by style transferring, GANs are presented. Finally, generated images are diagnosed by doctors and EyeNet is used to check the performance. Furthermore, observation of similarity among some diseases are analyzed and described.

\noindent\textbf{4.1 Dataset Collection}

Experimental images come from the Retina Image Bank (RIB) \cite{rib}. Retinal image collection contains three types of photography that are fluorescein angiography(FA), optical coherence tomography (OCT) and color fundus photography (CFP). FA are gray-scale images and CFP are colorful images. CFP and FA imaging are reliable for whole fundus, and used as our dataset.  

\noindent\textbf{4.2 Setup}

As discussed above, we use images with AMD for experiments. Images contain CFP and FA type, and present the symptom of drusen and GA. All DNNs were implemented in PyTorch, and we modified the publicly available PyTorch code for the neural network algorithm. Details of various methods are described later, respectively. The derivative of all generative models is sped in CUDA for gradient-based optimization.

\noindent\textbf{4.3 Style Transferring Neural Networks}

\begin{figure}[h]
  \begin{minipage}[t]{0.33\linewidth} 
    \centering 
    \includegraphics[width=1.5in]{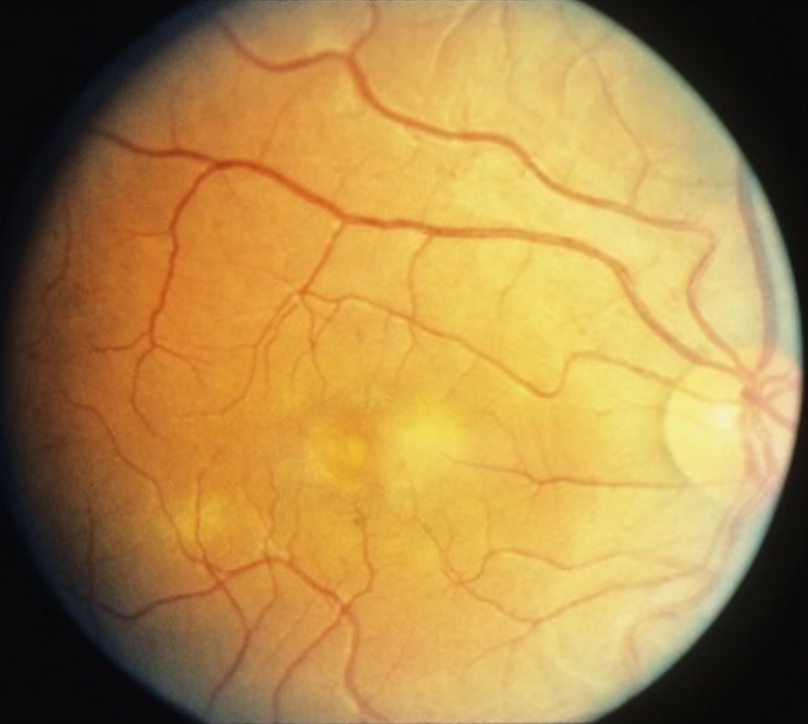}\\ 
    (a)
  \end{minipage}%
  \begin{minipage}[t]{0.33\linewidth} 
    \centering 
    \includegraphics[width=1.38in]{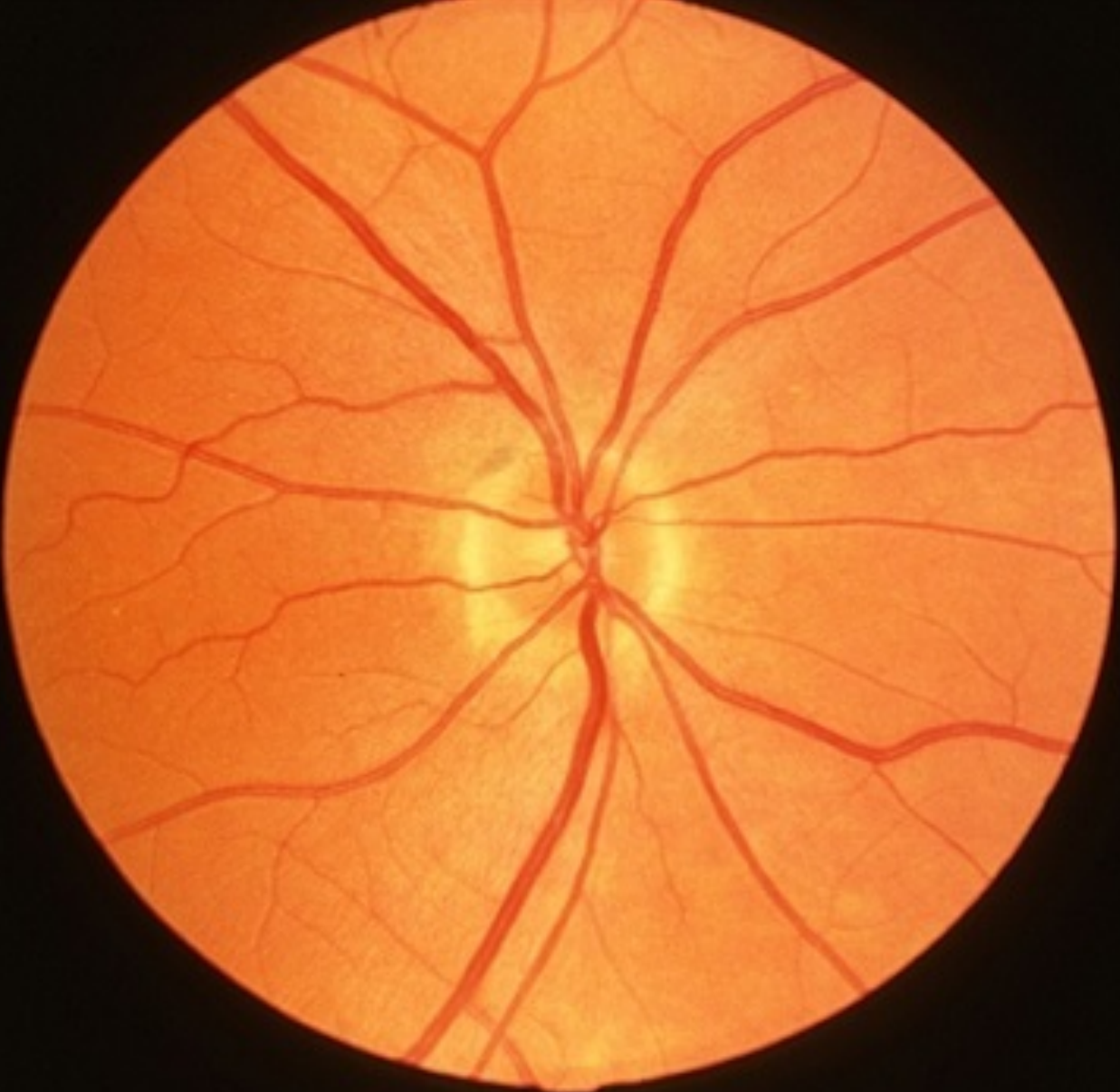} \\
    (b)
  \end{minipage}%
  \begin{minipage}[t]{0.3\linewidth} 
    \centering 
    \includegraphics[width=1.54in]{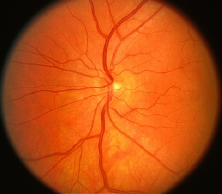}\\ 
    (c)
  \end{minipage}
  \caption{Three CFP fundus images that are used to generate new images are seen as style images.}
  \label{fig:figure2}
\end{figure}

\begin{figure}[!htb]
\centering
\subfigure[]{
\begin{minipage}[t]{0.33\linewidth}
\centering
\includegraphics[width=1.5in]{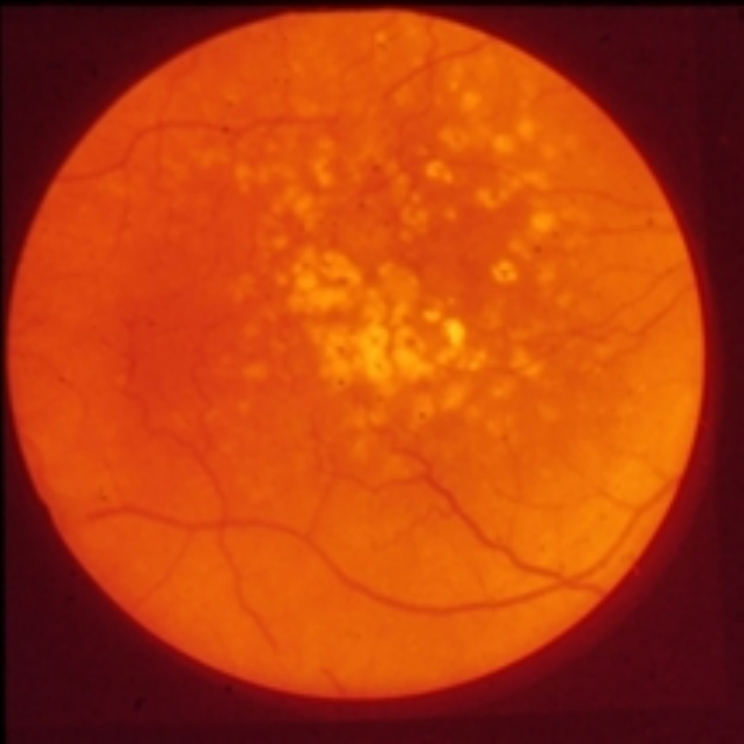}
\end{minipage}%
}%
\subfigure[]{
\begin{minipage}[t]{0.33\linewidth}
\centering
\includegraphics[width=1.5in]{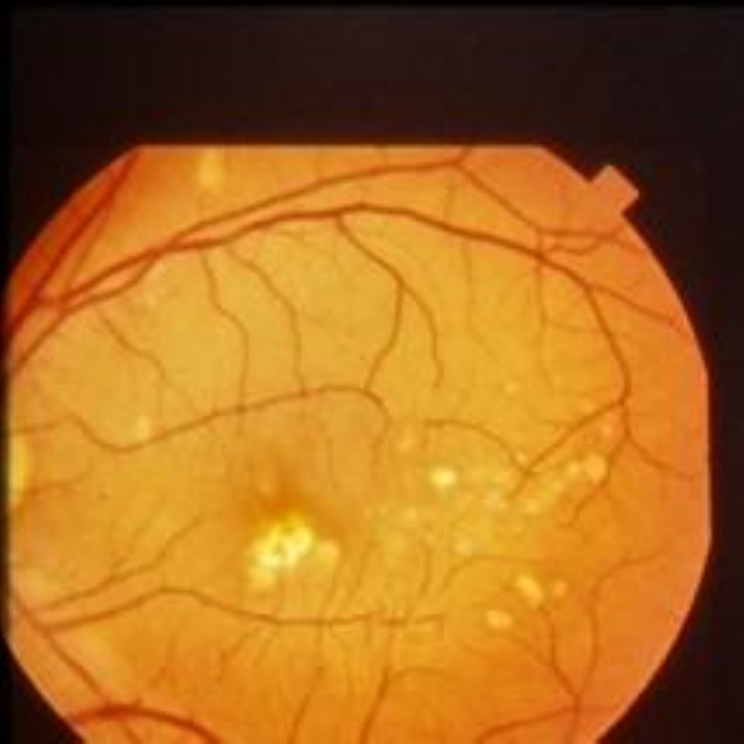}
\end{minipage}%
}%
\subfigure[]{
\begin{minipage}[t]{0.33\linewidth}
\centering
\includegraphics[width=1.5in]{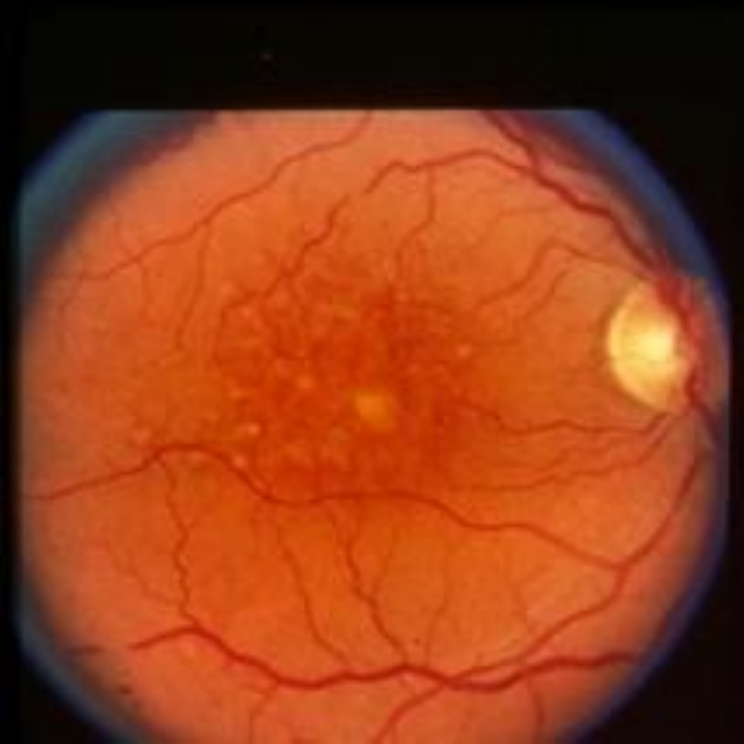}
\end{minipage}%
}%
       
\subfigure[]{
\begin{minipage}[t]{0.33\linewidth}
\centering
\includegraphics[width=1.5in]{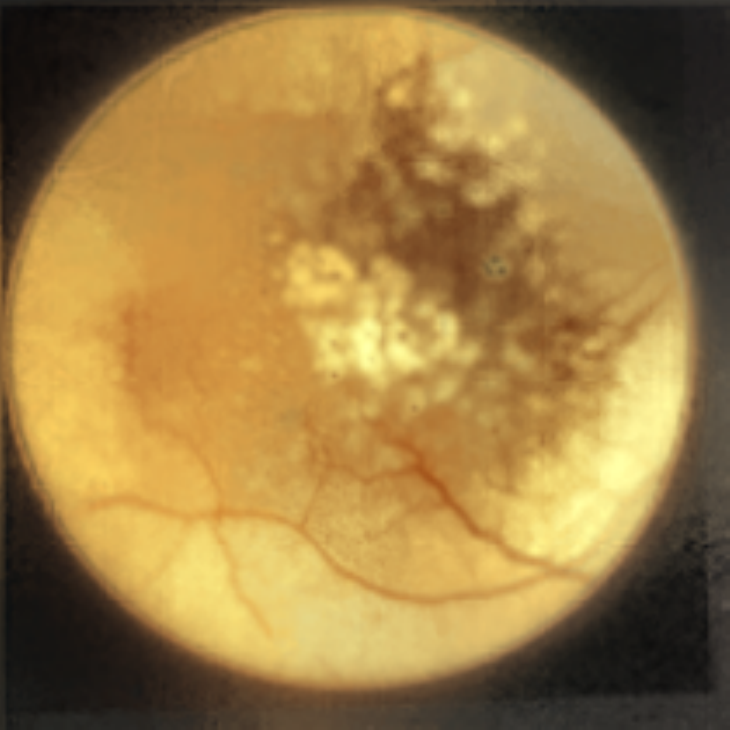}
\end{minipage}%
}%
\subfigure[]{
\begin{minipage}[t]{0.33\linewidth}
\centering
\includegraphics[width=1.5in]{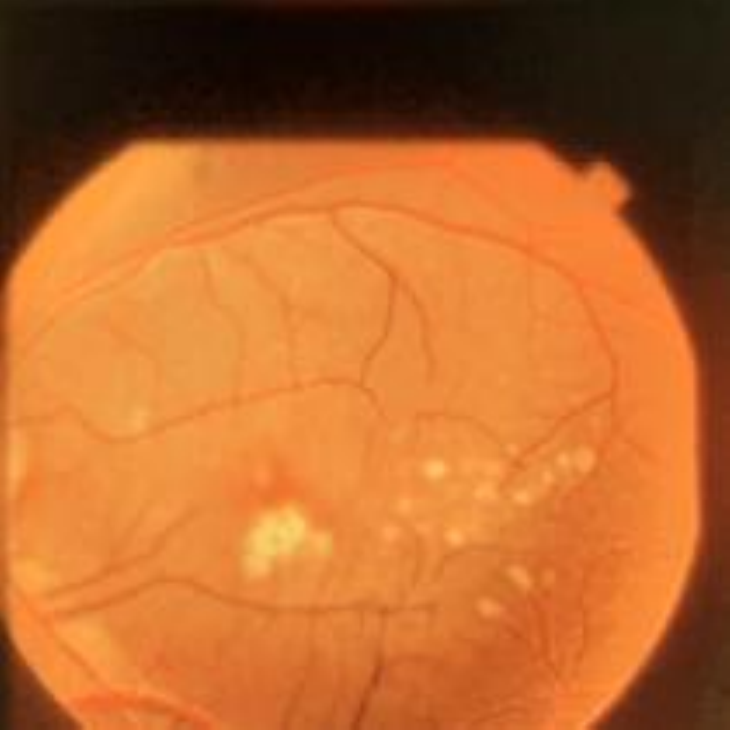}
\end{minipage}%
}%
\subfigure[]{
\begin{minipage}[t]{0.33\linewidth}
\centering
\includegraphics[width=1.5in]{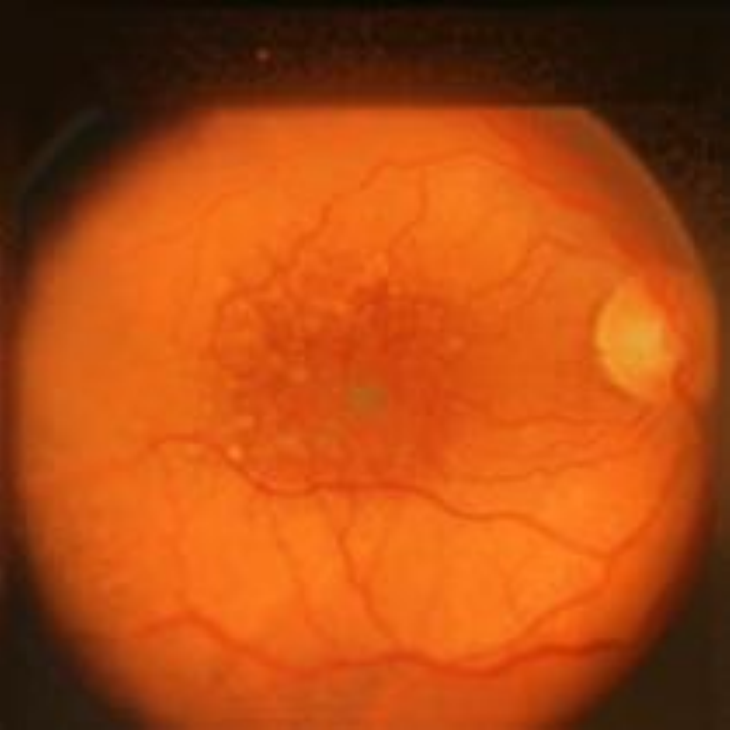}

\end{minipage}%
}%

\centering
\caption{ Three CFP fundus images with symptom of drusen and corresponding generated images. (a), (b), (c) Original images. (d), (e), (f) Generated images.}
\label{fig:figure3}
\end{figure}

\begin{figure}[h]
\centering
\subfigure[]{
\begin{minipage}[t]{0.33\linewidth}
\centering
\includegraphics[width=1.5in]{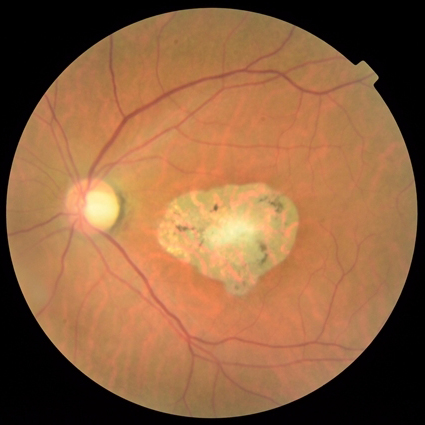}
\end{minipage}%
}%
\subfigure[]{
\begin{minipage}[t]{0.33\linewidth}
\centering
\includegraphics[width=1.5in]{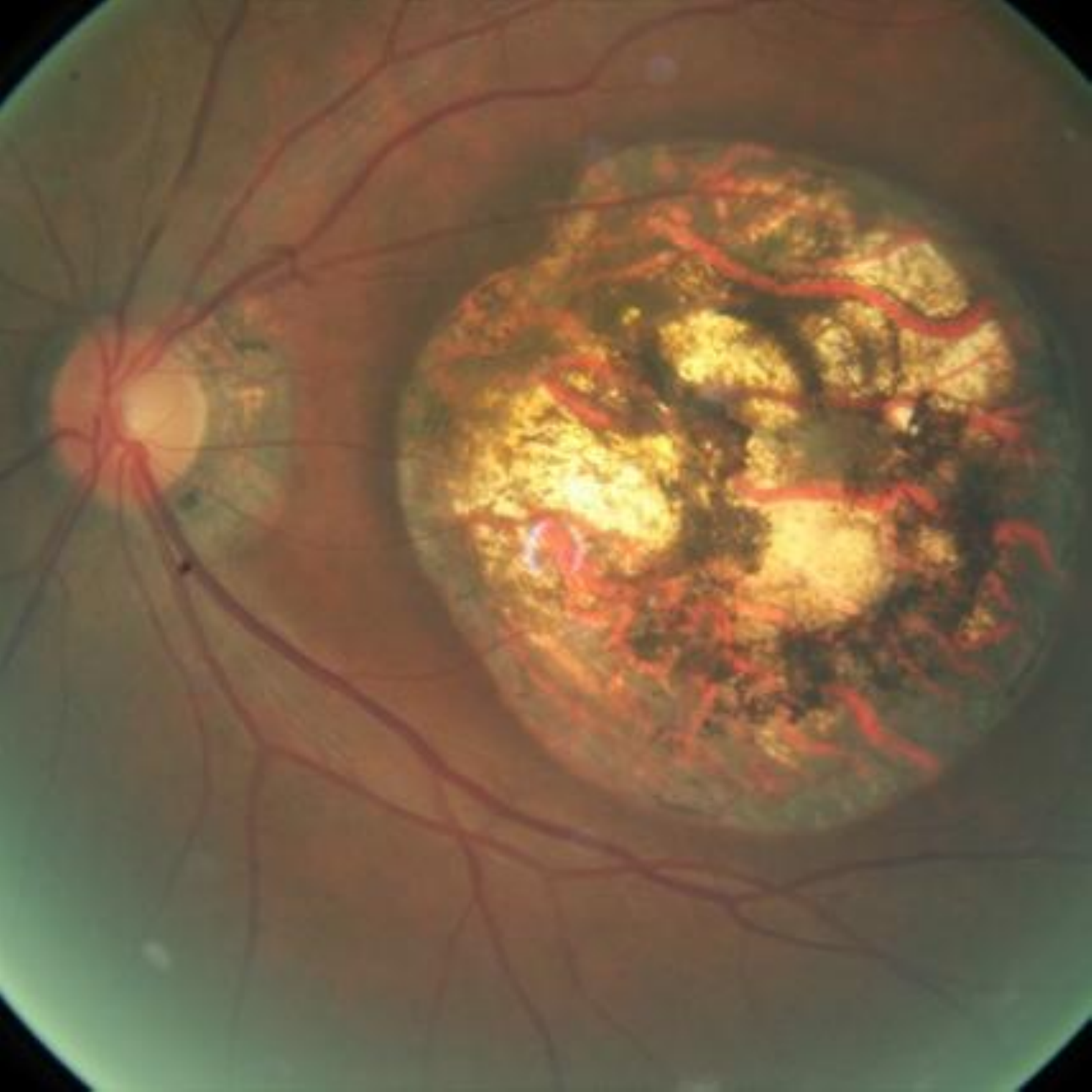}
\end{minipage}%
}%
\subfigure[]{
\begin{minipage}[t]{0.33\linewidth}
\centering
\includegraphics[width=1.5in]{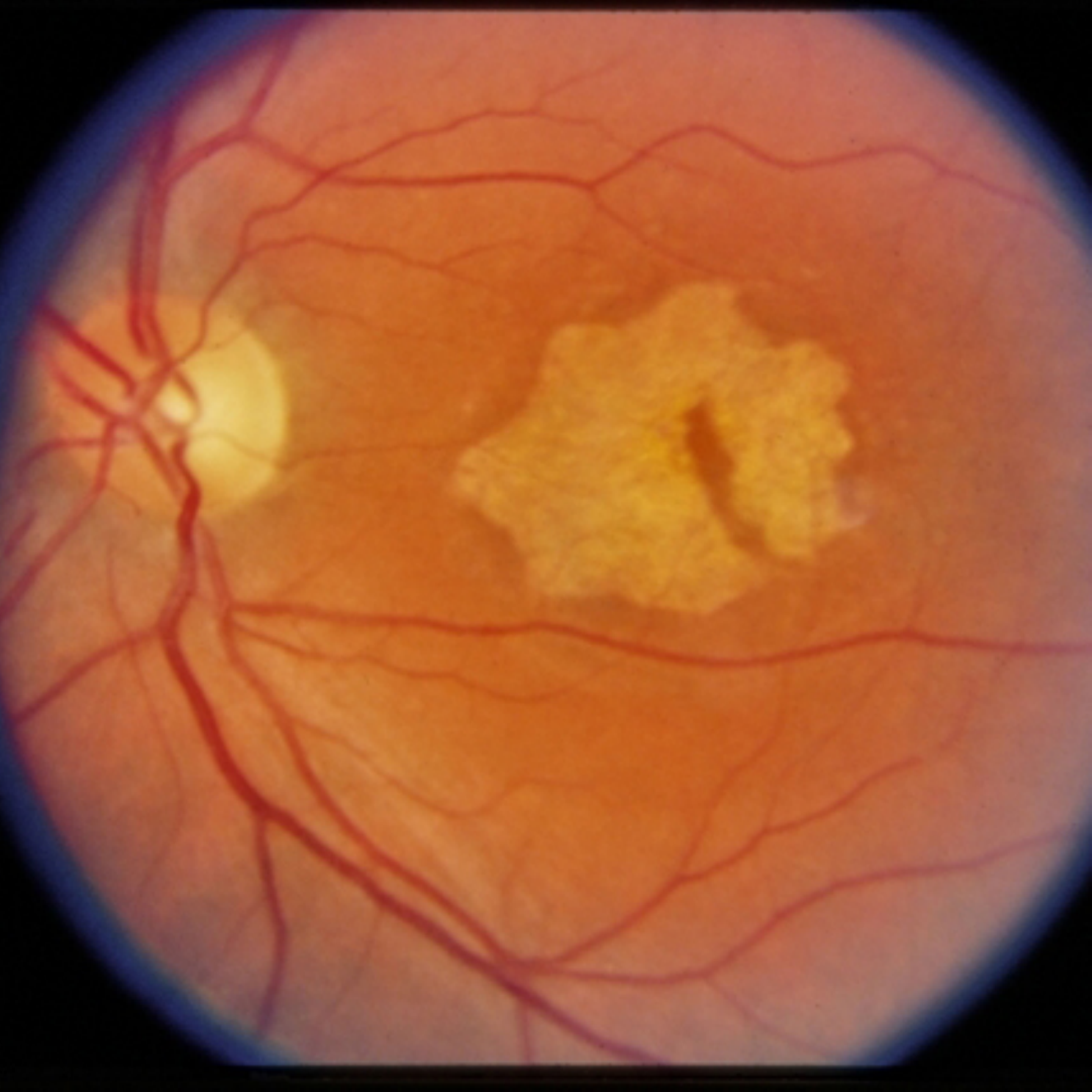}
\end{minipage}%
}%
       
\subfigure[]{
\begin{minipage}[t]{0.33\linewidth}
\centering
\includegraphics[width=1.5in]{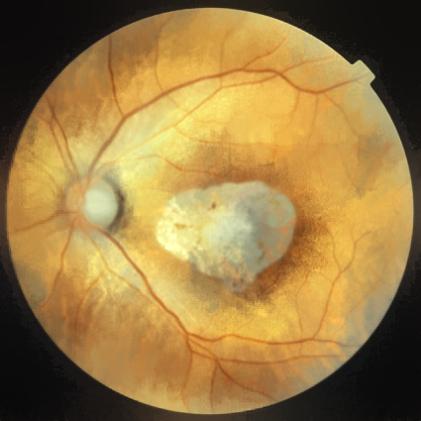}
\end{minipage}%
}%
\subfigure[]{
\begin{minipage}[t]{0.33\linewidth}
\centering
\includegraphics[width=1.5in]{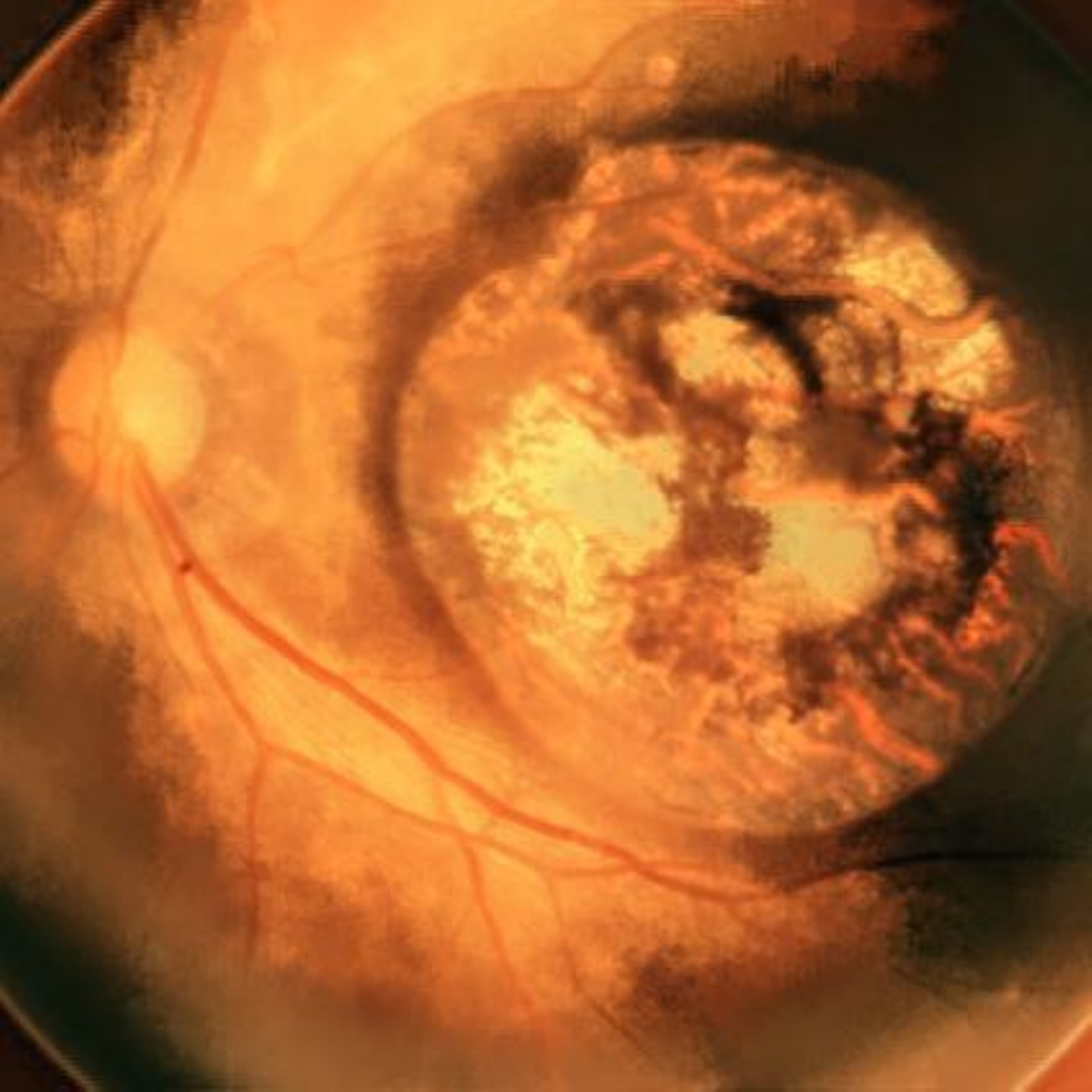}
\end{minipage}%
}%
\subfigure[]{
\begin{minipage}[t]{0.33\linewidth}
\centering
\includegraphics[width=1.5in]{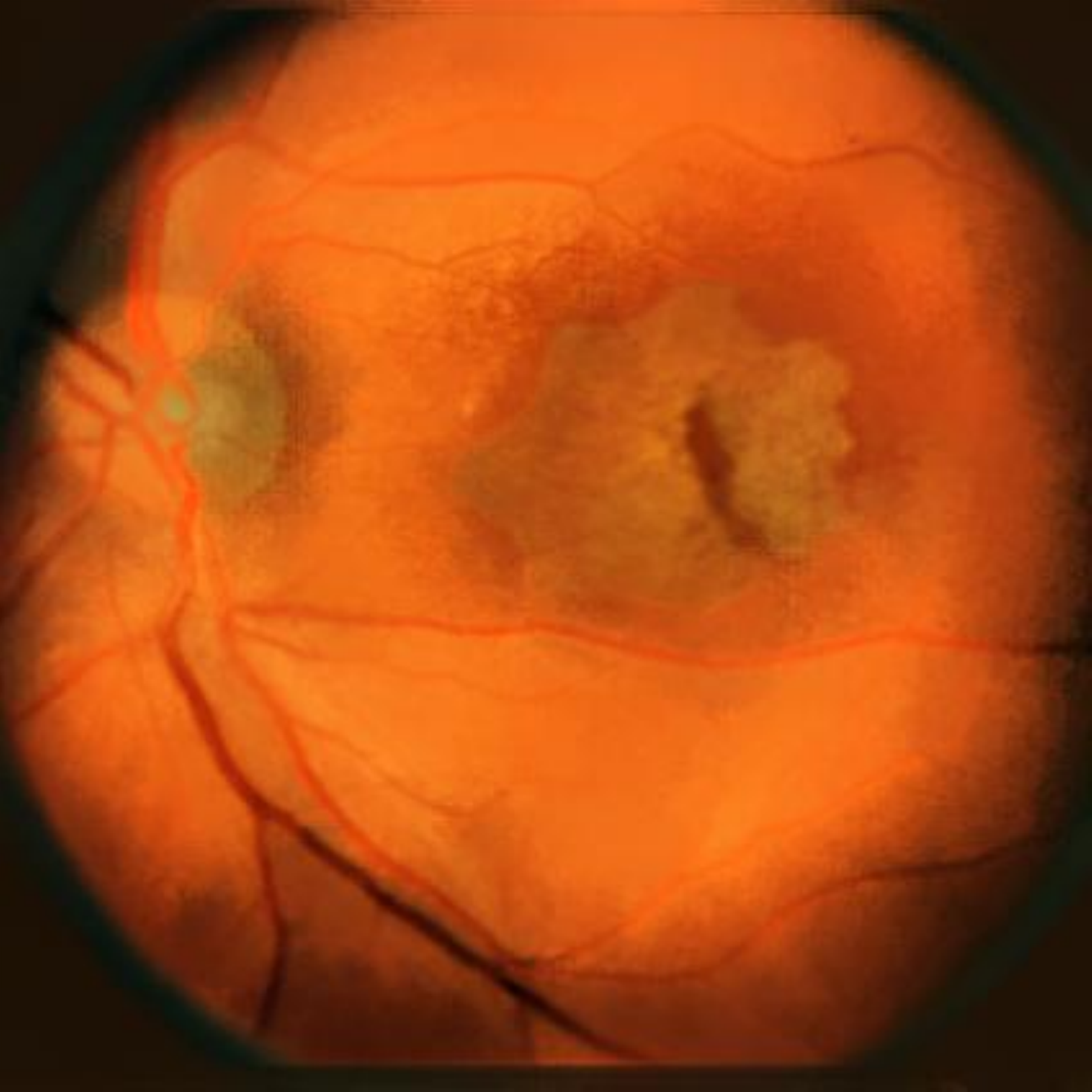}

\end{minipage}%
}%

\centering
\caption{ Three CFP fundus images with symptom of GA and corresponding generated images. (a), (b), (c) Original images. (d), (e), (f) Generated images.}
\label{fig:figure4}
\end{figure}

\begin{figure}[!htb]
  \begin{minipage}[t]{0.33\linewidth} 
    \centering 
    \includegraphics[width=1.5in]{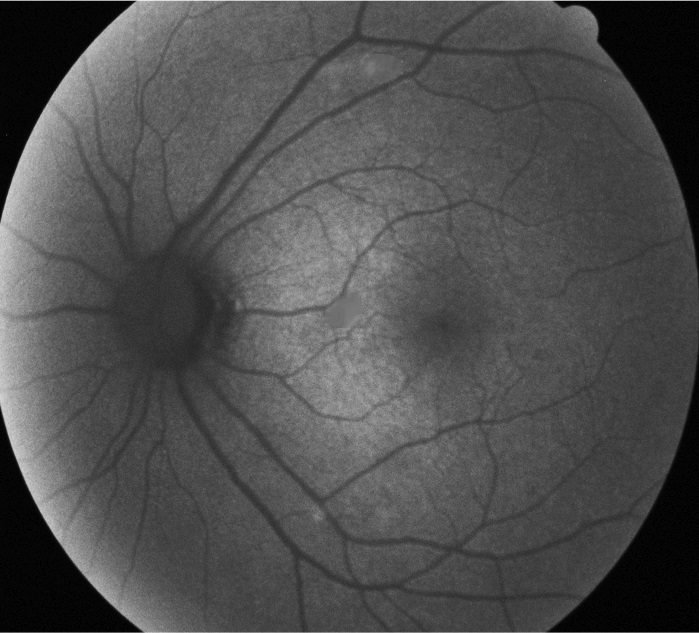}\\ 
    (a)
  \end{minipage}%
  \begin{minipage}[t]{0.33\linewidth} 
    \centering 
    \includegraphics[width=1.35in]{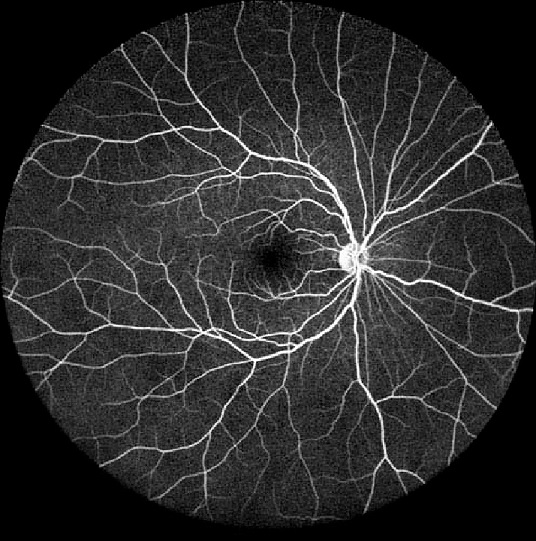} \\
    (b)
  \end{minipage}%
  \begin{minipage}[t]{0.34\linewidth} 
    \centering 
    \includegraphics[width=1.3in]{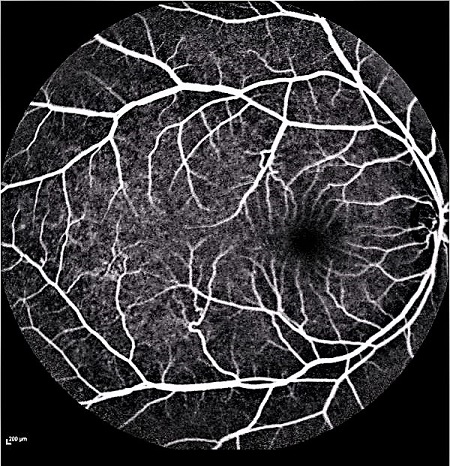}\\ 
    (c)
  \end{minipage}
  \caption{Three AF fundus images that are used to generate new images are seen as style images.}
  \label{fig:figure21}
\end{figure}

\begin{figure}[h]
\centering
\subfigure[]{
\begin{minipage}[t]{0.33\linewidth}
\centering
\includegraphics[width=1.5in]{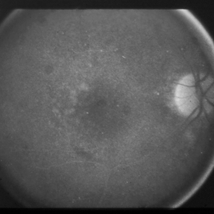}
\end{minipage}%
}%
\subfigure[]{
\begin{minipage}[t]{0.33\linewidth}
\centering
\includegraphics[width=1.5in]{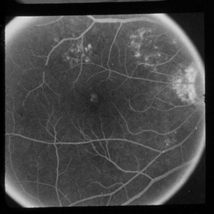}
\end{minipage}%
}%
\subfigure[]{
\begin{minipage}[t]{0.33\linewidth}
\centering
\includegraphics[width=1.5in]{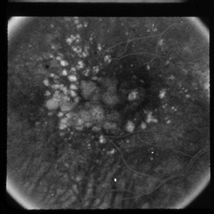}
\end{minipage}%
}%
       
\subfigure[]{
\begin{minipage}[t]{0.33\linewidth}
\centering
\includegraphics[width=1.5in]{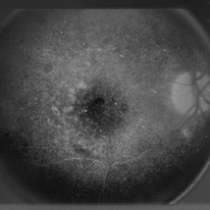}
\end{minipage}%
}%
\subfigure[]{
\begin{minipage}[t]{0.33\linewidth}
\centering
\includegraphics[width=1.5in]{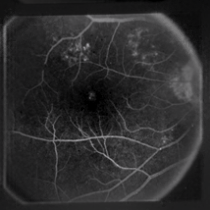}
\end{minipage}%
}%
\subfigure[]{
\begin{minipage}[t]{0.33\linewidth}
\centering
\includegraphics[width=1.5in]{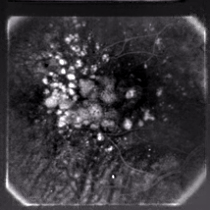}
\end{minipage}%
}%

\centering
\caption{ Three FA fundus images with symptom of drusen and corresponding generated images. (a), (b), (c) Original images. (d), (e), (f) Generated images.}
\label{fig:figure12}
\end{figure}

\begin{figure}[h]
\centering
\subfigure[]{
\begin{minipage}[t]{0.33\linewidth}
\centering
\includegraphics[width=1.5in]{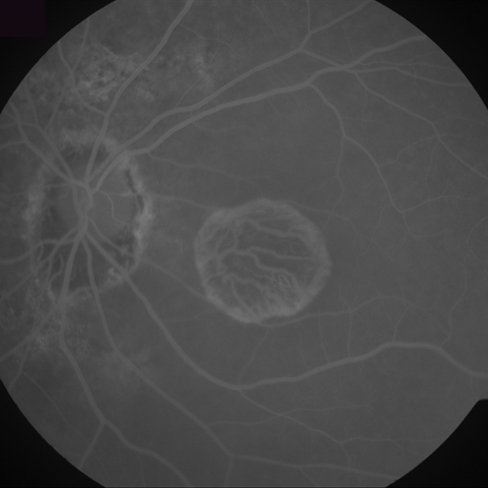}
\end{minipage}%
}%
\subfigure[]{
\begin{minipage}[t]{0.33\linewidth}
\centering
\includegraphics[width=1.5in]{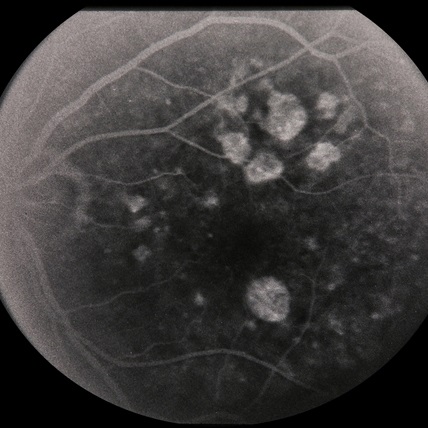}
\end{minipage}%
}%
\subfigure[]{
\begin{minipage}[t]{0.33\linewidth}
\centering
\includegraphics[width=1.5in]{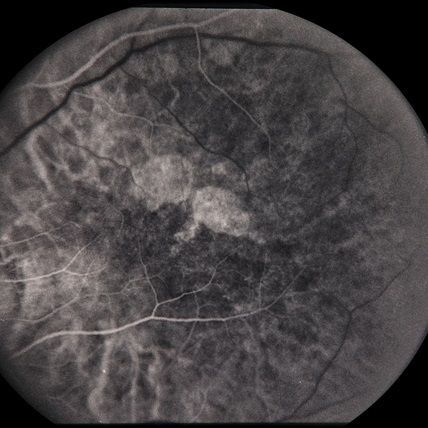}
\end{minipage}%
}%
       
\subfigure[]{
\begin{minipage}[t]{0.33\linewidth}
\centering
\includegraphics[width=1.5in]{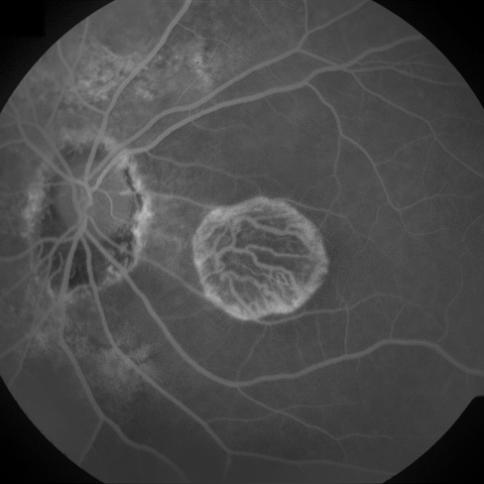}
\end{minipage}%
}%
\subfigure[]{
\begin{minipage}[t]{0.33\linewidth}
\centering
\includegraphics[width=1.5in]{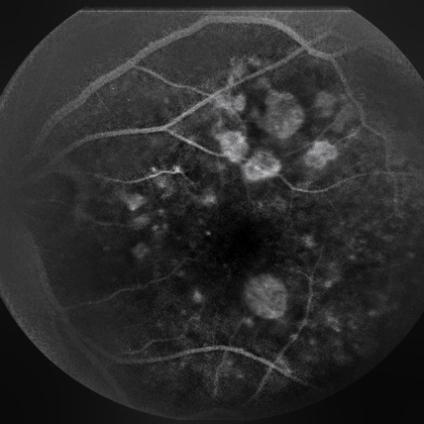}
\end{minipage}%
}%
\subfigure[]{
\begin{minipage}[t]{0.33\linewidth}
\centering
\includegraphics[width=1.5in]{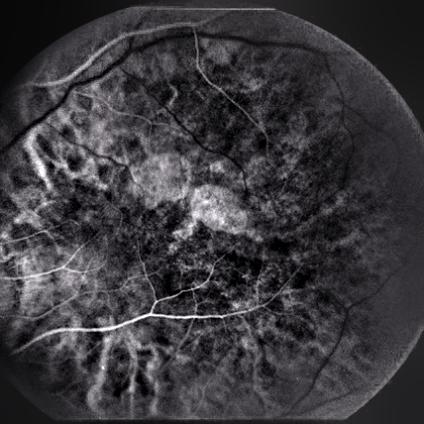}
\end{minipage}%
}%
\centering
\caption{ Three FA fundus images with symptom of GA and corresponding generated images. (a), (b), (c) Original images. (d), (e), (f) Generated images.}
\label{fig:figure13}
\end{figure}

Style transferring neural network in \cite{li2018closed} was modified to generate new disease images. This network adopts layers from "conv1\underline{ }1" to "conv4\underline{ }1" in pre-trained VGG-19 \cite{simonyan2014very} network for the encoder, whose weights are provided by ImageNet-pretrained weights. What's more, multi-level stylization strategy proposed in \cite{li2018closed} is applied to optimize the VGG features in different layers. Input images are three CFP images and three FA images as style images shown in Fig. \ref{fig:figure2} and \ref{fig:figure21}. Six CFP images with three drusen and three GA images in Fig. \ref{fig:figure3} and Fig. \ref{fig:figure4}. Also, FA images are applied to generate new images in Fig. \ref{fig:figure12} and Fig. \ref{fig:figure13}. For CFP images, six images are shown in Fig. \ref{fig:figure3} and in Fig. \ref{fig:figure4}. In Fig. \ref{fig:figure3}, generated images contain round, discrete yellow-white dots, which are the symptom of drusen. In the same way, in Fig. \ref{fig:figure4}, well-demarcated areas appear on the three images. Therefore, style transferring can generate new retinal symptom images.


Furthermore, generated images from FA images are presented in Fig. \ref{fig:figure12} and Fig. \ref{fig:figure13}. Results in the images are nearly identical to the original images, because original networks are applied to stylize color images. However, six generated images contain more concise features than the original ones, which helps ophthalmologists make better judgments. Therefore, this style transferring networks can fulfill edge sharpening and enhancement of contrast. No matter which kinds of images are generated, advanced features in new disease images still exist. Furthermore, analyses of image performance by EyeNet and CAMs for prediction are presented in a later section. 

\noindent \textbf{4.4 DCGANs and WGANs}

\begin{figure}[t]
  \begin{minipage}[t]{0.5\linewidth} 
    \centering 
    \includegraphics[width=1.5in]{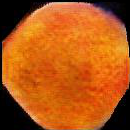}\\ 
    (a)
  \end{minipage}%
  \begin{minipage}[t]{0.5\linewidth} 
    \centering 
    \includegraphics[width=1.5in]{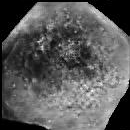}\\ 
    (b)
  \end{minipage}
  \caption{Drusen images generated by DCGANs.(a) CFP image. (b) FA image.}
  \label{fig:figure5}
\end{figure}

\begin{figure}[!htb]
\centering
\subfigure[]{
\begin{minipage}[t]{0.5\linewidth}
\centering
\includegraphics[width=1.5in]{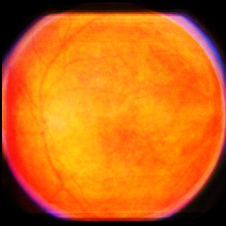}
\end{minipage}%
}%
\subfigure[]{
\begin{minipage}[t]{0.5\linewidth}
\centering
\includegraphics[width=1.5in]{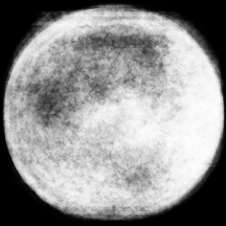}
\end{minipage}%
}%

\subfigure[]{
\begin{minipage}[t]{0.5\linewidth}
\centering
\includegraphics[width=1.5in]{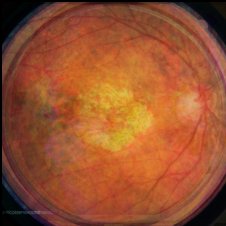}
\end{minipage}%
}%
\subfigure[]{
\begin{minipage}[t]{0.5\linewidth}
\centering
\includegraphics[width=1.5in]{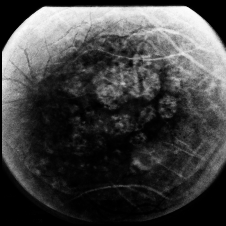}
\end{minipage}%
}%

\label{fig:figure20}
\centering
\caption{ Drusen and GA images generated by WGANs.(a) Generated drusen CFP image. (b) Generated drusen FA image. (c) Generated drusen CFP image. (d) Generated GA FA image.}
\end{figure}

In this section, DCGANs and WGANs are trained with thousands of CFP and FA images that have symptoms of drusen and GA separately; both of the models require four to six hours to train. Generated images have been diagnosed by ophthalmologists for verification. Images generated by DCGANs, which are shown in Fig. \ref{fig:figure5}, cannot be identified as a valid retinal image with symptom. However, drusen and GA images generated by WGANs can be used by ophthalmologists to diagnose. In Fig. \ref{fig:figure20}, generated drusen images are diagnosed as insignificant of drusen but can be identified by EyeNet. As for generated GA images in Fig. \ref{fig:figure20}, irregularly shaped macular atrophy can be identified by an ophthalmologist. Macular atrophy is a distinguishable trait of GA, which means WGANs indeed learn the symptoms of drusen and GA from specific AMD and generate new images. Thus, WGANs perform better than DCGANs because of resolution. Structure of DCGANs limits the size of generated images to be 64x64, so some pathological details are lost. We choose WGANs for following experiments.

\noindent\textbf{4.5 EyeNet Results for prediction}

EyeNet in \cite{yang2018novel} is trained to predict the accuracy of images to accomplish the pipeline. Though ImageNet and our retinal dataset are much different, using pre-trained weights on ImageNet rather than random ones has boosted testing accuracy of any models with 5 to 15 percent. Besides, pre-trained models tend to converge much faster than random initialized. The training images encompass 52 kinds of fundus images, which are randomly divided into three parts: 70\% for training, 10\% for validation and 20\% for testing. It is noted that synthesized images are not used to train this network. The training lasts 400 epochs. The first 200 epochs take a learning rate of 1e-4 and the second 200 take 1e-5. Besides, we apply random data augmentation during training. In every epoch, $70\%$ probability for a training sample is affinely transformed.  After EyeNet is trained, generated images are fed into it, and the average predicted probabilities are shown in Table 1. Compared to drusen, The accuracy declines when it comes to identifying generated geographic atrophy images. The lack of geographic atrophy images in the EyeNet dataset weaken the capability of the classifier to discern traits about geographic atrophy. Despite of the setback, there is a exciting exploration that the predictions are not randomly distributed but focus on particular diseases, which is likely caused by the high-dimensional features mentioned above.

\begin{table}[!htb]
  \centering
    \caption{{Average accuracy classified by EyeNet.}}
\begin{tabular}{|c|c|c|c|c|ll}

\cline{1-5}
              & Drusen-CFP & Drusen-FA & GA-CFP & GA-FA & & \\[4pt] \cline{1-5}
 Real Images  & 0.442  & 0.368 & 0.166 & 0.128 & &\\[3pt] \cline{1-5}
 WGANs        & \textbf{0.594} & 0.374 & 0.254 & 0.291 & &\\[3pt] \cline{1-5}
 Style Transferring & 0.376  & \textbf{0.657} & 0.139 & 0.231 & &\\[3pt] \cline{1-5}
\end{tabular}
\label{table:table1}
\end{table}

\noindent\textbf{4.6 Image Sample Size Effect }

\begin{figure*}[t]
  \centering
\includegraphics[width=\linewidth,scale=0.6]{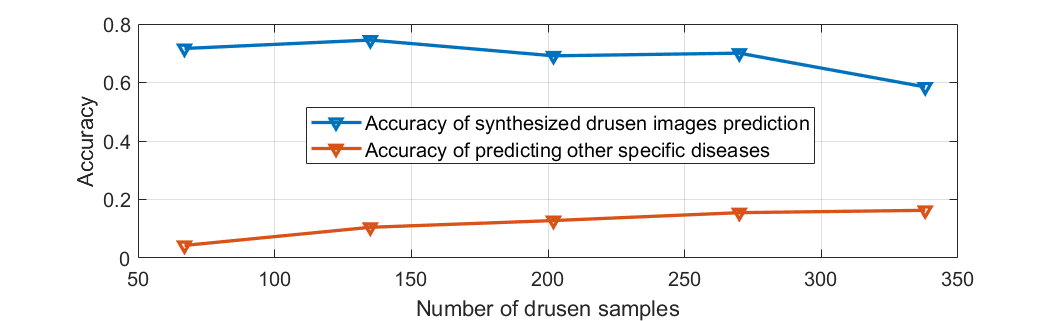}
    \caption{{Size effect of image samples and probability of predicting other specific diseases.}}
\label{fig:figure18}
\end{figure*}

Generative models are data-driven and the performance highly depending on the sample size. The EyeNet dataset we use contains 19496 retinal images with 1448 AMD images. In this section, We choose 338 drusen images as samples to test the size effect of GANs. Experiments show the difficulty of synthesizing high quality images rises along with the increase of the sample number. Fig. \ref{fig:figure18} shows accuracy of successfully predicting synthesized images, but AMD slightly declines as the sample number increases. In general, the more samples used to train a generative, the harder it is to extract specific visual features for generative model, which requires images with similarity; this is difficult to achieve when it comes to biological traits. On the other hand, prediction error focuses on some specific diseases, and the probability of predicting these diseases rises when the sample number falls. The phenomenon implies that high dimensional features in the retinal images exist. Furthermore, with more sample images, we can more likely to detect the symptom. This is a pathological approach to reveal hidden relations among diseases. 

\begin{figure*}[!hbt]
  \centering
\includegraphics[width=0.8\linewidth,scale=0.8]{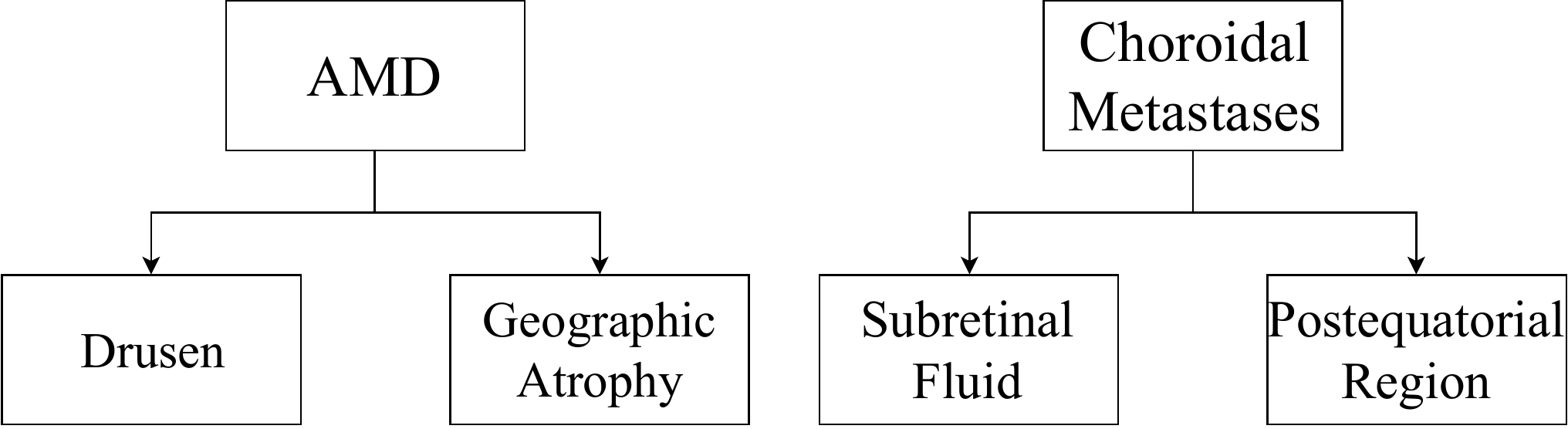}
    \caption{{Clinical hierarchical relationship of AMD and Choroidal metastases from \cite{COLEMAN20081835} and \cite{mewis1982breast}. Drusen increase a person’s risk of developing AMD; Geographic atrophy (GA) is an advanced form of age-related macular degeneration that can result in the progressive and irreversible loss of retina. Subretinal fluid and equatorial retina are significant features in choroidal metastases from clinical understanding. }}
\label{fig:figure17}
\end{figure*}

\begin{figure*}[!hbt]
  \centering
\includegraphics[width=0.8\linewidth,scale=0.8]{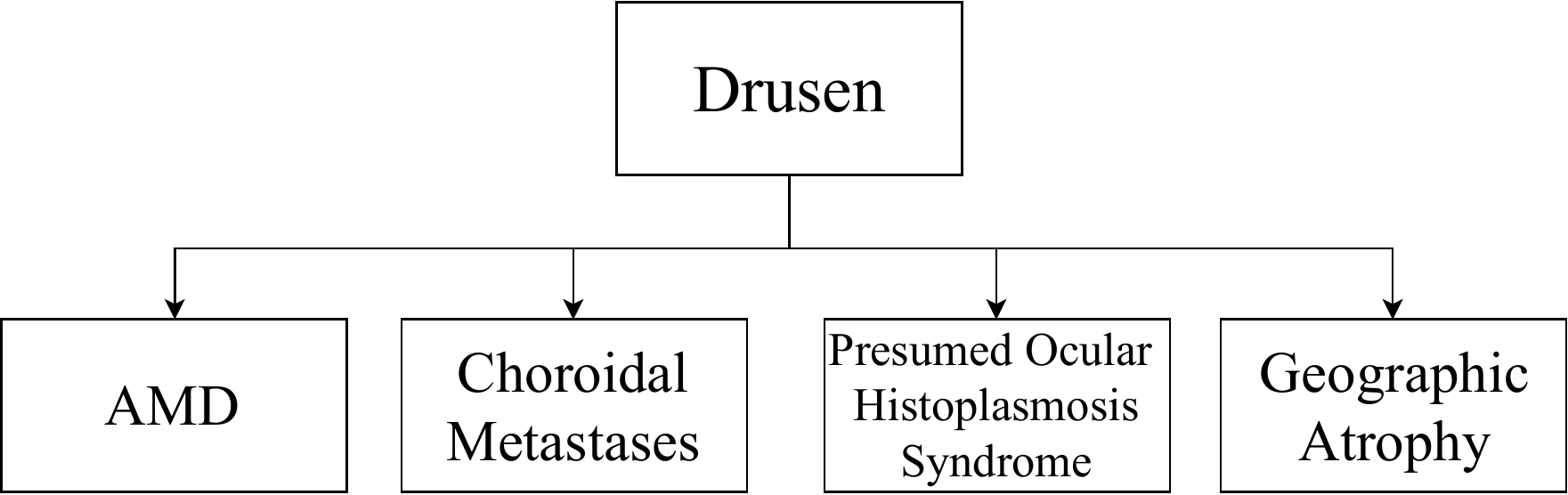}
    \caption{{Pathological Classification referring to GANs results. As a visual representation learned by deep network, Drusen are correlated to AMD (58.3\%), Choroidal metastasis (5.12\%), Presumed-ocular-histoplasmosis-syndrome (6.52\%), and Pattern dystrophy (2.57\%). }}
\label{fig:figure16}
\end{figure*}

\noindent\textbf{4.7 Pathological retinal diseases classification inspired by size effect of GANs}
 With higher quality images and thriving computer vision skills, visible retinal disease symptoms can being detected and represented. Based on traditional classification, symptoms have pathological correlations among retinal diseases for ophthalmologist to use in diagnoses, as shown in Fig. \ref{fig:figure17}. However, according to the discovery above, retinal diseases have hidden relation connected by invisible features. With GANs, we can propose a method to improve current classification. In this case, the classification could modified by the results of GANs as shown in Fig. \ref{fig:figure16}.


\noindent\textbf{4.8 Neural Network Visualization for Retinal Images} 

\begin{figure}[!hbt]
\centering
\subfigure[]{
\begin{minipage}[t]{0.33\linewidth}
\centering
\includegraphics[width=1.55in]{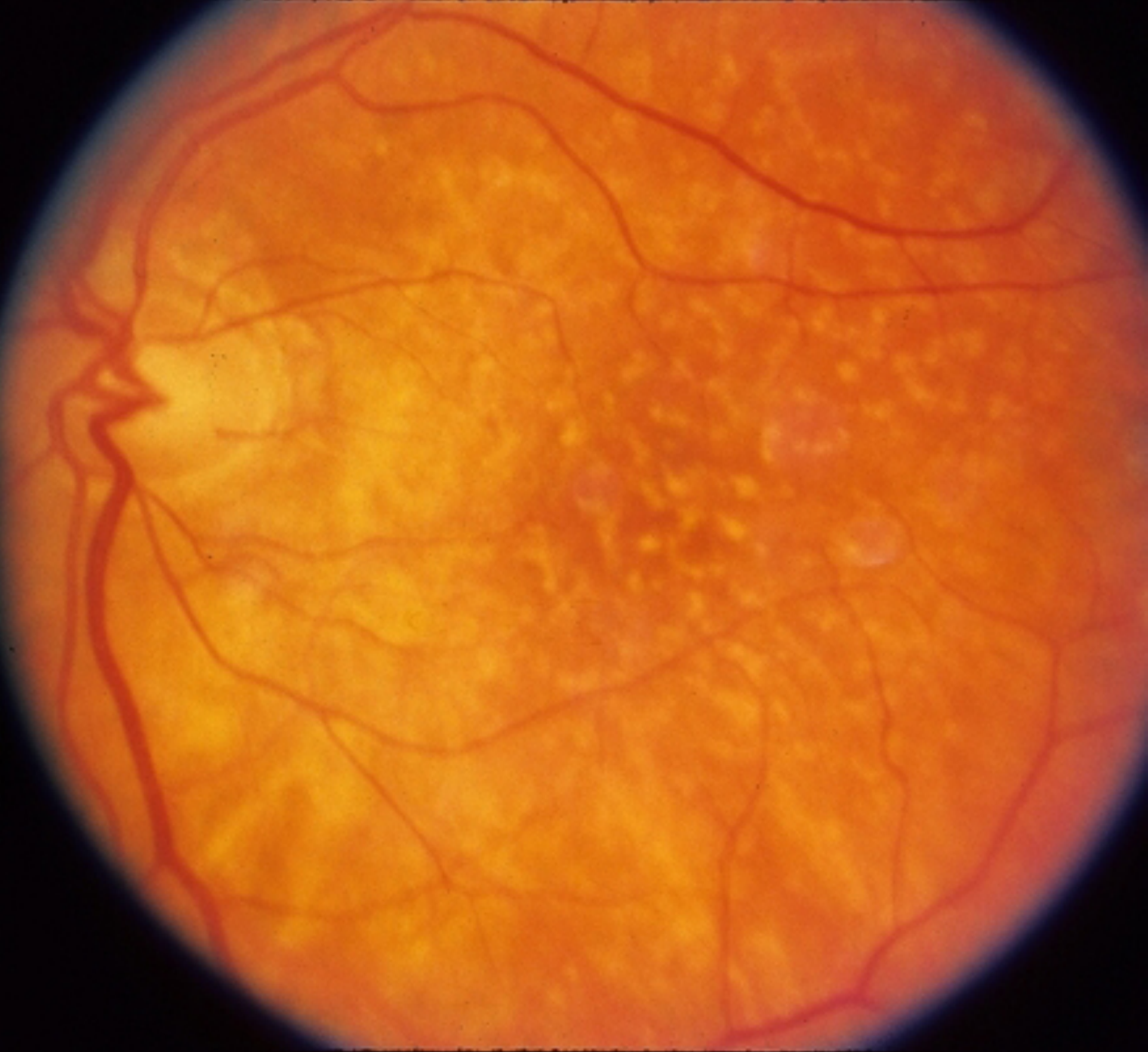}
\end{minipage}%
}%
\subfigure[]{
\begin{minipage}[t]{0.33\linewidth}
\centering
\includegraphics[width=1.425in]{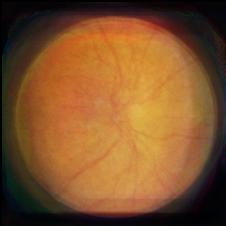}
\end{minipage}%
}%
\subfigure[]{
\begin{minipage}[t]{0.33\linewidth}
\centering
\includegraphics[width=1.425in]{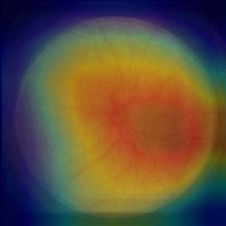}
\end{minipage}%
}%
       
\subfigure[]{
\begin{minipage}[t]{0.33\linewidth}
\centering
\includegraphics[width=1.5in]{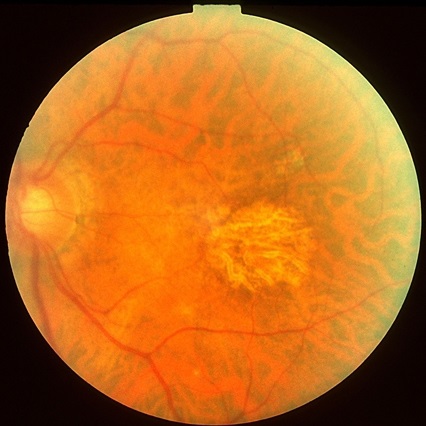}\\ 
\end{minipage}%
}%
\subfigure[]{
\begin{minipage}[t]{0.33\linewidth}
\centering
\includegraphics[width=1.5in]{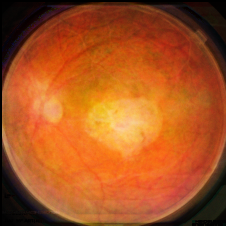}
\end{minipage}%
}%
\subfigure[]{
\begin{minipage}[t]{0.33\linewidth}
\centering
    \includegraphics[width=1.5in]{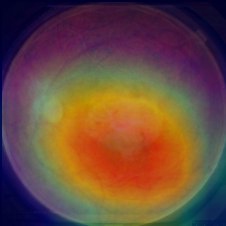} 

\end{minipage}%
}%

\subfigure[]{
\begin{minipage}[t]{0.33\linewidth}
\centering
\includegraphics[width=1.5in]{image_style_transferring/fig_7_a.jpeg}
\end{minipage}%
}%
\subfigure[]{
\begin{minipage}[t]{0.33\linewidth}
\centering
\includegraphics[width=1.5in]{image_style_transferring/fig_7_d.jpeg} 
\end{minipage}%
}%
\subfigure[]{
\begin{minipage}[t]{0.33\linewidth}
\centering
\includegraphics[width=1.5in]{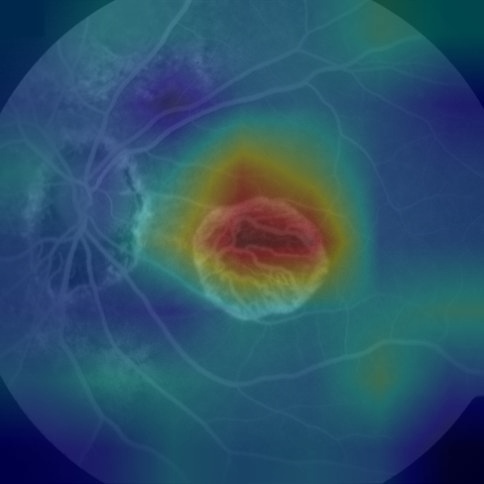}
\end{minipage}%
}%
\centering
\caption{ Generated images and their CAMs. (a) Original drusen image (CFP). (b) Generated image of (a) by WGANs. (c) CAM of (b). (d) Original GA image (CFP). (e) Generated image of (d) by WGANs. (f) CAM of (e). (g) Original GA image (FA). (h) Generated image of (g) by style trnasferring. (i) CAM of (g).}
\label{fig:figure19}
\end{figure}

Finally, we verified the hypothesis that vessel-based segmentation and contrast enhancement are two coherent features to decide the type of retinal diseases. Using techniques of generating CAMs introduced in \cite{zhou2015cnnlocalization}, we visualized feature maps of the final convolutional layer of ResNet50 in Fig. \ref{fig:figure19}. In our results, generated drusen images are well identified. However, generated GA images are not focused on the exact location of the symptom, but they are close. As discussed above, in the clinical diagnosis process, "vessel patterns" and "fundus structure" are the most crucial features for identifying the symptoms of different diseases. These types of features cover more than 80\% of retinal diseases \cite{crick2003textbook,akram2005common}.

\section{Conclusions and Future Work}

We have implemented style transferring, DCGANs and WGANs to generate disease images that are detailed to capture different stages of AMD. Symptoms of images are drusen and GA; both FA and CFP images are generated. Images from DCGANs are difficult to be identified due to limit of resolution. However, images from style transferring and WGANs are easier to identify by ophthalmologists,and generated images preserve pathological details. EyeNet is used to predict the disease label, and results of generated drusen images are similar to original images. However, generated GA images are more distant compared to original images, because of the small number of GA images used during training EyeNet. This phenomenon shows that generated new images can be fed into the classifier to improve it. Also, CAMs are useful for extracting label-specific features. In Fig. \ref{fig:figure19}(c),(f) and(i), warmer color parts are located in the well-demarcated areas or spots, which represents disease features that are close to those parts.


 \par In this paper, only a small number of disease images are synthesized and evaluated, so various images can be tested and enhanced further. Furthermore, different kinds of skills like semantic segmentation can be merged into the original GANs framework. With better and diverse generated images, classifier can be trained robustly and applied to predict the disease more precisely. Above all, a re-trained network discovers hidden relationships and provides ophthalmologists with useful disease features warranting further investigation.

\bibliographystyle{splncs}
\bibliography{egbib}

\end{document}